\documentclass[10pt,twocolumn,letterpaper]{article}

\usepackage{iccv}
\usepackage{times}
\usepackage{epsfig}
\usepackage{graphicx}
\usepackage{amsmath}
\usepackage{amssymb}
\usepackage{balance}
\usepackage{float}
\usepackage{multirow}
\usepackage{amsfonts}
\usepackage{color}
\usepackage{colortbl}
\definecolor{mygray}{gray}{.96}
\usepackage{authblk}

\DeclareMathOperator{\softmax}{softmax}
\DeclareMathOperator{\slsu}{S-LSU}
\DeclareMathOperator{\tlsu}{T-LSU}
\DeclareMathOperator{\stls}{ST-LS_{block}}
\DeclareMathOperator{\conv}{Conv}
\DeclareMathOperator{\uniform}{Uniform}
\DeclareMathOperator{\Softmax}{Softmax}

\usepackage[pagebackref=true,breaklinks=true,letterpaper=true,colorlinks,bookmarks=false]{hyperref}

\iccvfinalcopy 

\def\iccvPaperID{****} 

\begin{document}

\title{UNIK: A Unified Framework for Real-world Skeleton-based Action Recognition}

\renewcommand{\thefootnote}{\fnsymbol{footnote}}

\author{\vspace{-0.3cm}
Di Yang\textsuperscript{1,2*}\hskip 1em
Yaohui Wang\textsuperscript{1,2*}\hskip 1em
Antitza Dantcheva\textsuperscript{1,2}\hskip 1em
Lorenzo Garattoni\textsuperscript{3} \hskip 1em

Gianpiero Francesca\textsuperscript{3}\hskip 1em
François Brémond\textsuperscript{1,2}
\\
\textsuperscript{1}Inria \hskip 1em
\textsuperscript{2}Université Côte d'Azur \hskip 1em
\textsuperscript{3}Toyota Motor Europe \\


{\tt\small \{di.yang, yaohui.wang, antitza.dantcheva, francois.bremond\}@inria.fr} \hskip 1em

{\tt\small \{lorenzo.garattoni, gianpiero.francesca\}@toyota-europe.com}

}

\renewcommand\Authands{ and }

\ificcvfinal\thispagestyle{empty}\fi

\twocolumn[{%
\maketitle
\vspace{-0.9cm}
\begin{figure}[H]
\hsize=\textwidth 
\centering
\includegraphics[width=1.0\textwidth ]{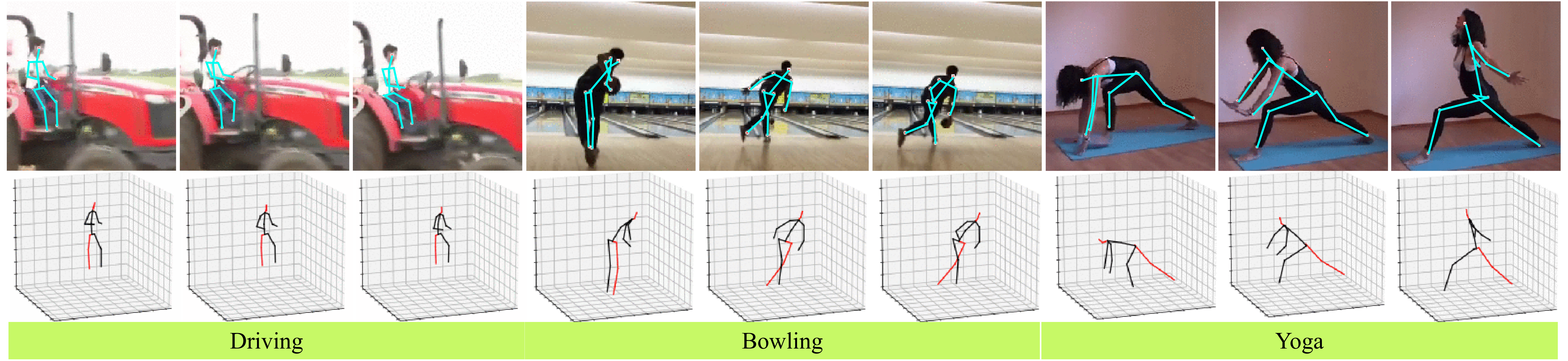}
\caption{\textbf{Skeleton-based action recognition.} In this paper we introduce a generic skeleton-based action recognition model \textbf{UNIK}, and report significant improvements by using our model pre-trained on \textbf{Posetics}, a real-world, large, human skeleton video dataset.}
\label{figure:intro}
\end{figure}
}]

\begin{abstract}

\vspace{-0.2cm}
Action recognition based on skeleton data has recently witnessed increasing attention and progress. State-of-the-art approaches adopting Graph Convolutional networks (GCNs) can effectively extract features on human skeletons relying on the pre-defined human topology. Despite associated progress, GCN-based methods have difficulties to generalize across domains, especially with different human topological structures. In this context, we introduce UNIK\footnote{Equal contribution. \\~~~~Code is available at: \url{https://github.com/YangDi666/UNIK}}, a novel skeleton-based action recognition method that is not only effective to learn spatio-temporal features on human skeleton sequences but also able to generalize across datasets. This is achieved by learning an optimal dependency matrix from the uniform distribution based on a multi-head attention mechanism.
Subsequently, to study the cross-domain generalizability of skeleton-based action recognition in real-world videos, we re-evaluate state-of-the-art approaches as well as the proposed UNIK in light of a novel Posetics dataset. This dataset is created from Kinetics-400 videos by estimating, refining and filtering poses. We provide an analysis on how much performance improves on smaller benchmark datasets after pre-training on Posetics for the action classification task. Experimental results show that the proposed UNIK, with pre-training on Posetics, generalizes well and outperforms state-of-the-art when transferred onto four target action classification datasets: Toyota Smarthome, Penn Action, NTU-RGB+D 60 and NTU-RGB+D 120. 

\vspace{-0.2cm}
\end{abstract}

\section{Introduction}\label{intro}
\begin{figure}[t]
\begin{center}

\includegraphics[width=0.95\linewidth]{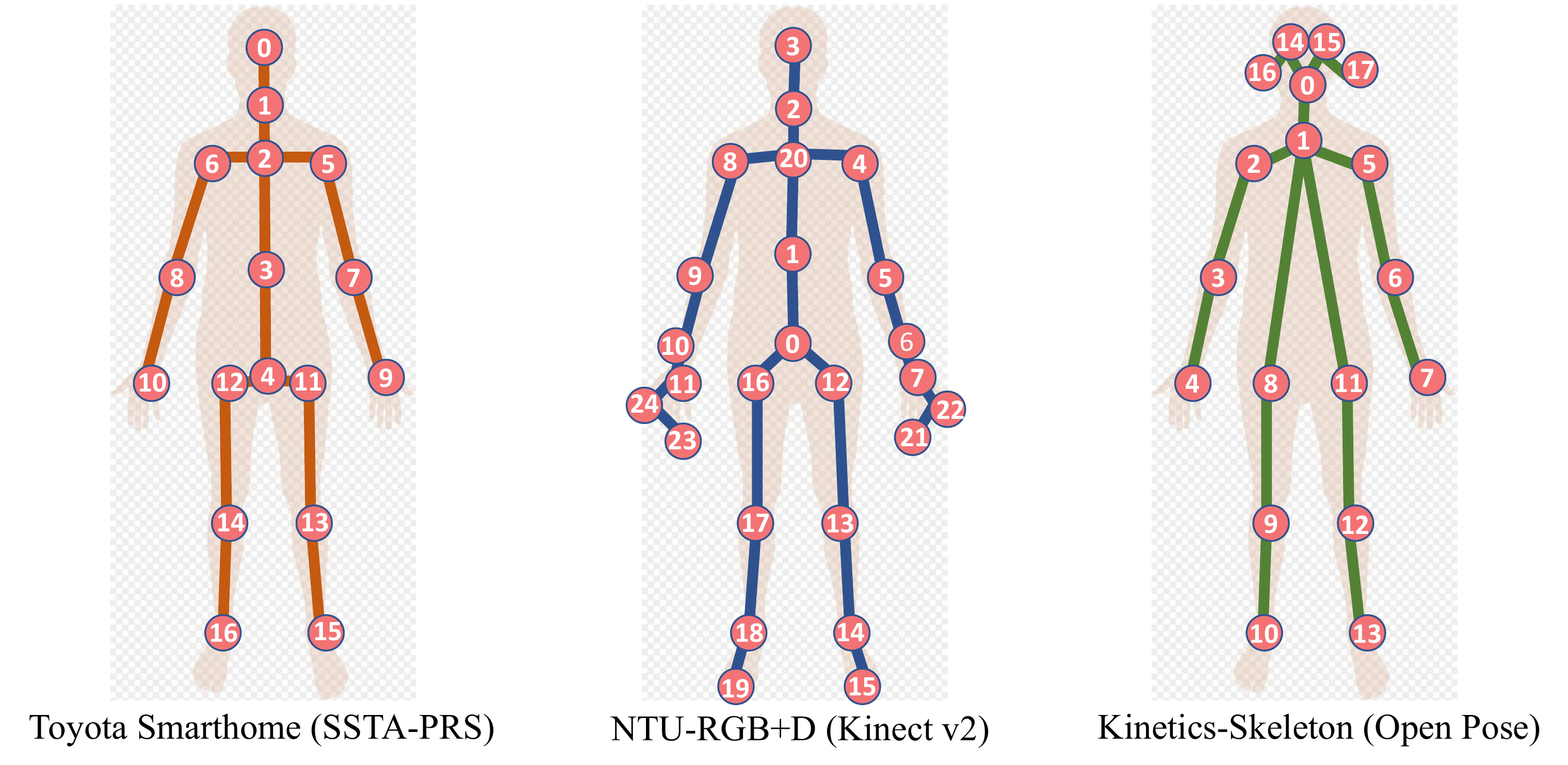}
\end{center}
\vspace{-0.5cm}
\caption{\text{Human joint labels} of three datasets: Toyota Smarthome (left), NTU-RGB+D (middle) and Kinetics-Skeleton (right). We note the different numbers, orders and locations of joints.}
\vspace{-0.5cm}
\label{fig:joints}
\end{figure}

As skeleton-based human action recognition methods rely on 2D or 3D positions of human key joints only, they are able to filter out noise caused, for instance, by background clutter, changing light conditions, and to focus on the action being performed~
%
\cite{Vemulapalli2014HumanAR, zhang2019view, Tanfous2019SparseCO, 8784961, Xie2018MemoryAN, Li2018CooccurrenceFL, Caetano2019SkeletonIR, Yan2018SpatialTG, Li_2019_CVPR, 2sagcn2019cvpr,Gao2019OptimizedSA, shi_skeleton-based_2019, Peng2020LearningGC, Shi2019SkeletonBasedAR, liu2020disentangling}. Recent approach, namely Graph Convolutional Networks (GCNs)~\cite{Yan2018SpatialTG}, models human joints, as well as their natural connections (\ie, bones) in skeleton spatio-temporal graphs to carry 
both, spatial and temporal inferences. Consequently, several successors, applying Adaptive GCNs (AGCNs), with optimized graph construction strategies to extract multi-scale structural features on skeletons and long-range dependencies between joints have been proposed and shown encouraging results. Promising examples are graph convolutions
with learnable \textit{adjacency matrix}~\cite{2sagcn2019cvpr}, higher-order polynomials of \textit{adjacency matrix}~\cite{Li_2019_CVPR} and separate multi-scale subsets of \textit{adjacency matrix}~\cite{liu2020disentangling}. All these \textit{adjacency matrices} are manually pre-defined to represent the relationships between joints according to human topology.
Nevertheless, compared with RGB-based methods such as spatio-temporal Convolutional Neural Networks (CNNs)~\cite{Carreira_2017_CVPR,hara3dcnns} that are pre-trained on Kinetics~\cite{Carreira_2017_CVPR} to boost accuracy in downstream datasets and tasks, GCN-based models are limited as they are always trained individually on the target dataset (often small) from scratch. 
Our insight is that, the generalization abilities of these approaches are hindered by the need for different adaptive adjacency matrices when different topological human structures are used (\eg, joints number, joints order, bones), as in the case of the three datasets of Fig.~\ref{fig:joints}.
%
However, we note that such adaptive sparse \text{adjacency matrices} are transformed to fully dense matrices in deeper layers in order to capture long-range dependencies between joints. This new structure contradicts the initial and original topological skeleton structure.\par

Based on these considerations and as the human-intrinsic graph representation is deeply modified during training, we hypothesize that there should be a more optimized and generic initialization strategy that can replace the \textit{adjacency matrix}. To validate this hypothesis, we introduce UNIK, a novel unified framework for skeleton-based action recognition. In UNIK, the \textit{adjacency matrix} is initialized into a uniformly distributed \textit{dependency matrix} where each element represents the dependency weight between the corresponding pair of joints.
Subsequently, a multi-head aggregation is performed to learn and aggregate multiple dependency matrices by different attention maps. This mechanism jointly leverages information from several representation sub-spaces at different positions of the \textit{dependency matrix} to effectively learn the spatio-temporal features on skeletons. The proposed UNIK does not rely on any topology related to the human skeleton, 
makes it much easier to transfer onto other skeleton datasets. This opens up a great design space to further improve the recognition performance by transferring a model pre-trained on a sufficiently large dataset.
 \par

%

In addition, another difficulty for model generalization is that many skeleton datasets have been captured in lab environments with RGBD sensors (\eg, NTU-RGB+D~\cite{Shahroudy2016NTURA, NTU-120}). Then, the action recognition accuracy significantly decreases
, when the pre-trained models on the sensor data are transferred to the real-world videos, where skeleton data encounters a number of occlusions and truncations of the body. To address this, 
we create Posetics dataset by estimating and refining poses, as well as filtering, purifying and categorizing videos and annotations based on the real-world Kinetics-400~\cite{Carreira_2017_CVPR} dataset. To this aim, we apply multi-expert pose estimators~\cite{OpenPose, fang2017rmpe, RogezWS18} and a refinement algorithm~\cite{Yang_2021_WACV}. 
Our experimental analysis confirms: pre-training on Posetics improves state-of-the-art skeleton-based action recognition methods, when transferred and fine-tuned on all evaluated datasets~\cite{Das_2019_ICCV, penn, Shahroudy2016NTURA, NTU-120}.

In summary, the contributions of this paper are:
\begin{enumerate}\setlength{\itemsep}{0pt}
\item 
We go beyond GCN-based architectures by proposing UNIK with a novel design strategy by adopting dependency matrices and a multi-head attention mechanism for skeleton-based action recognition. 
%
%
%
%
\item 
We revisit real-world skeleton-based action recognition focusing on cross-domain transfer learning. The study is conducted on four target datasets with pre-training on Posetics, a novel and large-scale action classification dataset that features higher quality skeleton detections based on Kinetics-400.
\item 
We demonstrate that pre-training UNIK on Posetics and fine-tuning it on the target real-world datasets (\eg, Toyota Smarthome~\cite{Das_2019_ICCV} and Penn Action~\cite{penn}) can be a generic and effective methodology for skeleton-based action classification.



\end{enumerate}


\section{Related Work}

\paragraph{Skeleton-Based Action Recognition.}
Early skeleton-based approaches using Recurrent Neural Networks (RNNs)~\cite{zhang2019view, Tanfous2019SparseCO, 8784961, Song2017AnES, Xie2018MemoryAN} or Temporal Convolutional Networks (TCNs)~\cite{tcn} were proposed due to their high representation capacity. However, these approaches ignore the spatial semantic connectivity of the human body. Subsequently,~\cite{Li2018CooccurrenceFL, Caetano2019SkeletonIR, zhang2019view} proposed to map the skeleton as a pseudo-image (\ie, in a 2D grid structure to represent the spatial-temporal features) based on manually designed transformation rules and to leverage 2D CNNs to process the spatio-temporal local dependencies within the skeleton sequence by considering a partial human-intrinsic connectivity. ST-GCN~\cite{Yan2018SpatialTG} used spatial graph convolutions along with interleaving temporal convolutions for skeleton-based action recognition. This work considered the topology of the human skeleton, however ignored the important long-range dependencies between the joints. In contrast, recent AGCN-based approaches~\cite{Li_2019_CVPR, 2sagcn2019cvpr, Gao2019OptimizedSA, Shi2019SkeletonBasedAR, Peng2020LearningGC, shi_skeleton-based_2019, liu2020disentangling} have seen significant performance boost, by the advantage of improving the representation of human skeleton topology to process long-range dependencies for action recognition. Specifically,
2s-AGCN~\cite{2sagcn2019cvpr} introduced an adaptive graph convolutional network to adaptively learn the topology of the graph with self-attention, which was shown beneficial in action recognition.
Associated extension, MS-AAGCN~\cite{shi_skeleton-based_2019} incorporated multi-stream adaptive graph convolutional networks that used attention modules and 4-stream ensemble based on 2s-AGCN~\cite{2sagcn2019cvpr}.  
These approaches primarily focused on spatial modeling. Consequently, MS-G3D Net~\cite{liu2020disentangling} presented a unified approach for capturing complex joint correlations directly across space and time. However, the accuracy depends on the scale of the temporal segments, which should be carefully tuned for different datasets, preventing transfer learning.
Thus, previous approaches~\cite{2sagcn2019cvpr, shi_skeleton-based_2019, liu2020disentangling} learn adaptive adjacency matrices from the sub-optimal initialized human topology. In contrast, our work proposes an optimized and unified \textit{dependency matrix} that can be learned from the \textit{uniform distribution} by a multi-head attention process for skeleton-based action recognition without the constraint of human topology and a limited number of attention maps in order to improve performance, as well as generalization capacity.

\vspace{-0.3cm}

\paragraph{Model Generalization.} Previous methods~\cite{Yan2018SpatialTG, 2sagcn2019cvpr, shi_skeleton-based_2019, liu2020disentangling} were only evaluated on the target datasets, trained from scratch without taking advantages of fine-tuning on a pre-trained model. To explore the transfer ability for action recognition using human skeleton, recent research~\cite{sun2020viewinvariant, ConNTU} proposed view-invariant 2D or 3D pose embedding algorithms with pre-training performed on lab datasets~\cite{h36m_pami, NTU-120} that do not correspond to real-world and thus these techniques struggle to improve the action recognition performance on downstream tasks with large-scale real-world videos~\cite{Das_2019_ICCV, uav}. To the best of our knowledge, we are the first to explore the skeleton-based pre-training and fine-tuning strategies for real-world videos.


\section{Proposed Approach}



\subsection{Unified Architecture (UNIK)}
In this section we present UNIK, a unified spatio-temporal dependencies learning network for skeleton-based action recognition.
\vspace{-0.3cm}

\begin{figure*}
\begin{center}

\includegraphics[width=0.95\linewidth]{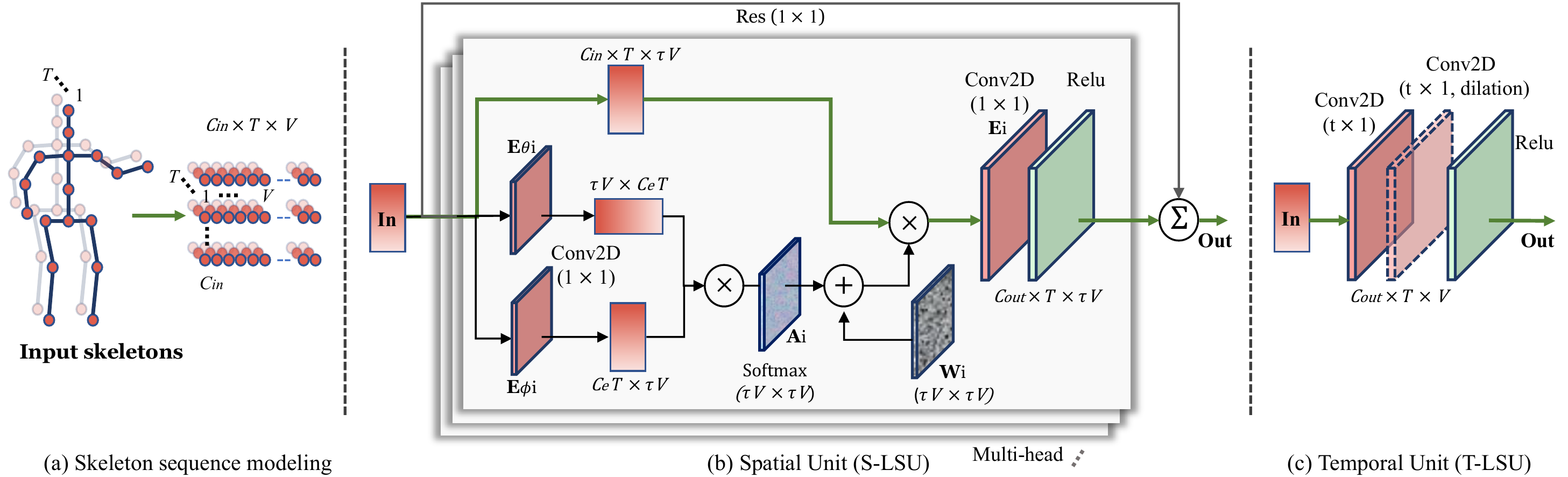}
\end{center}
   \vspace{-0.3cm}
   \caption{\small \textbf{Unified Spatial-temporal Network.} (a) The input skeleton sequence is modeled into a matrix with $C_{in}$ channels $\times$ $T$ frames $\times$ $V$ joints. (b) In each head of the S-LSU, the input data over a temporal sliding window ($\tau$) is multiplied by a dependency matrix which are obtained from the unified, uniformly initialized $\mathbf{W_i}$ and the self-attention based $\mathbf{A_i}$. $\mathbf{E}_i$ , $\mathbf{E}_{\theta i}$ and $\mathbf{E}_{\phi i}$ are for the channel embedding from $C_{in}$ to $C_{out}$/$C_{e}$ respectively by ($1\times1$) convolutions. The final output is the sum of the outputs from all the heads. (c) The T-LSU is composed of convolutional layers with ($t\times1$) kernels . $d$ denotes the dilation coefficient which can be different in each block.}
\vspace{-0.3cm}
\label{fig:comps}
\end{figure*}

\paragraph{Skeleton Sequence Modeling.}
As shown in Fig.~\ref{fig:comps} (a), the sequence of the input skeletons is modeled by a 3D spatio-temporal matrix, noted as $\mathbf{f_{in}}$. For each frame, the 2D or 3D body joint coordinates are arranged in a vector within the spatial dimension in any order as long as the order is consistent with other frames in the same video.
For the temporal dimension, the same body joints in two consecutive frames are connected. $T$, $V$, and $C_{in}$ represent the length of the video, the number of joints of the skeleton in one frame, as well as the input channels (2D or 3D at the beginning and expanded within the building blocks), respectively. The input $\mathbf{f_{in}}$ and the output $\mathbf{f_{out}}$ for each building block (see~\ref{p:overall}) are represented by a matrix in $\mathbb{R}^{C_{in}\times T\times V}$ and a matrix in $\mathbb{R}^{C_{out}\times T\times V}$, respectively.

\vspace{-0.3cm}

\begin{figure}[ht]

\begin{center}

\includegraphics[width=1.0\linewidth]{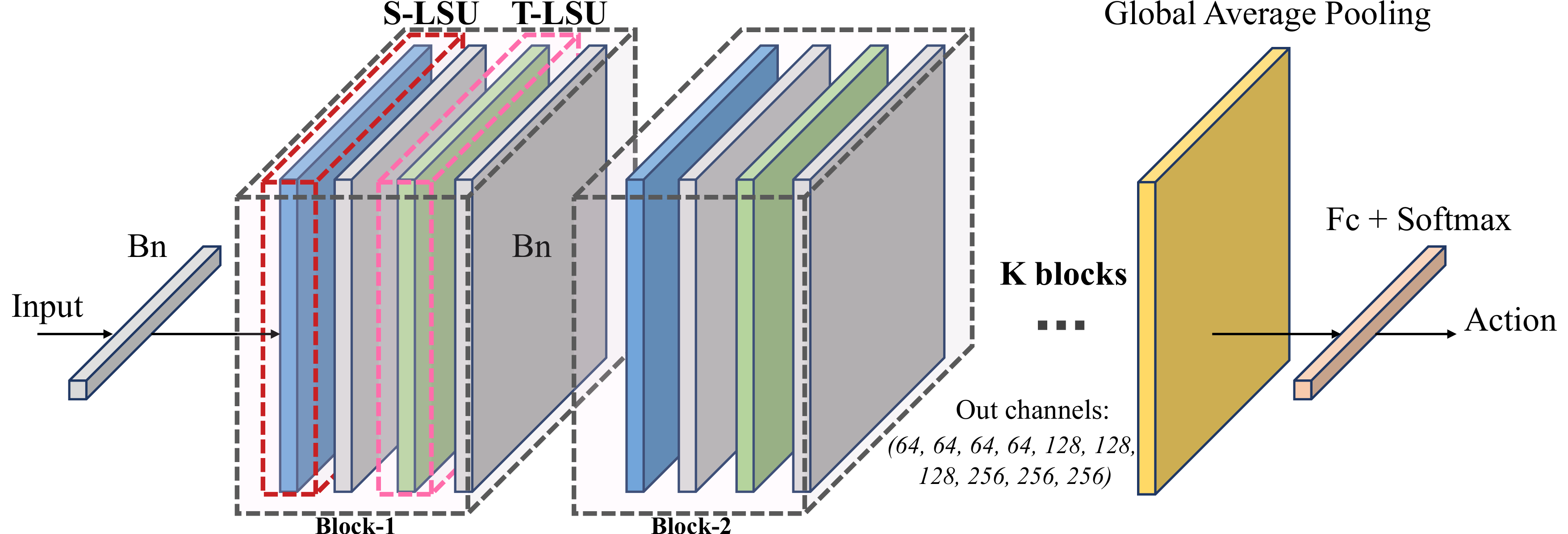}
\end{center}
   \vspace{-0.4cm}
   \caption{\small \textbf{Overall architecture.} There are K blocks with a 1D Batch normalization layer at the beginning, a global average pooling layer and a fully connected classifier at the end. Each block contains a Spatial Long-short dependency Unit \textbf{(S-LSU)}, a Temporal Long-short dependency Unit \textbf{(T-LSU)} and two Batch normalization layers.}
\vspace{-0.4cm}
\label{fig:architecture}
\end{figure}

\paragraph{Overall Architecture.}\label{p:overall} The overall architecture is composed of $K$ building blocks (see Fig.~\ref{fig:architecture}). Key components of each block constitute the Spatial Long-short Dependency learning Unit (S-LSU), as well as the Temporal Long-short Dependency learning Unit (T-LSU) that extract both spatial and temporal multi-scale features on skeletons over a large receptive field. The building block $\stls$ is formulated as follows:
\vspace{-0.1cm}
\begin{equation}\label{st-ldu}
    \mathbf{f_{out}}=\stls(\mathbf{f_{in}})=\tlsu\big(\slsu(\mathbf{f_{in}}) \big)
    \vspace{-0.1cm}
\end{equation}

S-LSU and T-LSU are followed by a 2D Batch normalization layer respectively. A 1D Batch normalization layer is added in the beginning for normalizing the flattened input data. Given a skeleton sequence, the modeled data is fed into the building blocks. After the last block, global average pooling is performed to pool feature maps of different samples to the same size. Finally, the fully connected classifier outputs the prediction of the human action. The number of blocks $K$ and the number of output channels should be adaptive to the size of the training set, as a large network cannot be trained with a small dataset. However, in this work, we do not need to adjust $K$, as we propose to pre-train the model on a large, generic dataset (see~\ref{sec:dataset}). We set $K=10$ with the number of output channels: $64, 64, 64, 64, 128, 128, 128, 256, 256, 256$ (see Fig.~\ref{fig:architecture}). In order to stabilize the training and ease the gradient propagation, a residual connection is added for each block.

\vspace{-0.3cm}

\paragraph{Spatial Long-short Dependency Unit (S-LSU).}\label{ld-learning} 

To aggregate the information from a larger spatial-temporal receptive field, a sliding temporal window of size $\tau$ is set over the input matrix. At each step, the input $\mathbf{f_{in}}$ across $\tau$ frames in the window becomes a matrix in $\mathbb{R}^{C_{in}\times T\times \tau V}$. For the purpose of spatial modeling, we use a multi-head and residual based S-LSU (see Fig.~\ref{fig:comps} (b)) and formulated as follows:
\vspace{-0.1cm}
\begin{equation}\label{s-ldu1}
    \mathbf{f_{out}}=\slsu(\mathbf{f_{in}})=\sum_{i=1}^N \mathbf{E}_i \cdot \big( \mathbf{f_{in}} \times (\mathbf{W}_i+\mathbf{A}_i) \big),
    \vspace{-0.1cm}
\end{equation}
where $N$ represents the number of heads. $\mathbf{E}_i\in \mathbb{R}^{C_{out}\times C_{in} \times 1 \times 1}$ denotes the 2D convolutional weight matrix with $1\times 1$ kernel size, which embeds the features from $C_{in}$ to $C_{out}$ by the dot product. $\mathbf{W}_i\in \mathbb{R}^{\tau V \times \tau V}$ is the ``dependency matrix" mentioned in Sec.~\ref{intro} to process the dependencies for every pair of spatial features. Note that $\mathbf{W}_i$ is learnable and uniformly initialized as random values within bounds (Eq.~\ref{s-ldu2}).
\vspace{-0.1cm}
\begin{equation}\label{s-ldu2}
    \mathbf{W}_{i}=\uniform(-bound, bound),
\end{equation}
where
\begin{equation}\label{s-ldu3}
    bound=\sqrt{\cfrac {6}{(1+a^2)V}},
    \vspace{-0.1cm}
\end{equation}
where $a$ denotes a constant indicating the negative slope~\cite{kaiming}. In this work, we take $a=\sqrt{5}$ as the initialization strategy of the fully connected layers, in order to efficiently find the optimized dependencies~\cite{kaiming}.
\vspace{-0.4cm}

\subparagraph{Self-attention Mechanism.} The matrix $\mathbf{A}_i$ in Eq.~\ref{s-ldu1} represents the non-local self attention map that adapts the dependency matrix $\mathbf{W}_i$ dynamically to the target action. This adaptive attention map is learned end-to-end with the action label. In more details, given the input feature map $\mathbf{f_{in}}\in \mathbb{R}^{C_{in}\times T \times \tau V}$, we first embed it into the space $\mathbb{R}^{C_e\times T\times \tau V}$ by two convolutional layers with $1\times1$ kernel size. The convolutional weights are denoted as $\mathbf{E}_{\theta i} \in \mathbb{R}^{C_{e}\times C_{in} \times 1 \times 1}$ and $\mathbf{E}_{\phi i} \in \mathbb{R}^{C_{e}\times C_{in} \times 1 \times 1}$, respectively. The two embedded feature maps are reshaped to $\tau V \times C_eT$ and $C_eT \times \tau V$ dimensions. They are then multiplied to obtain the attention map $\mathbf{A}_i \in \mathbb{R}^{\tau V \times \tau V}$, whose elements represent the attention weights between each two joints adapted to different actions. The value of the matrix is normalized to $0 \sim 1$ using a $\softmax$ function. We can formulate $\mathbf{A}_i$ as:

\begin{equation}\label{s-ldu4}
    \mathbf{A}_{i}=\Softmax \big( (\mathbf{E}^\mathbf{T}_{\theta i} \cdot \mathbf{f^T_{in}})\times(\mathbf{E}_{\phi i} \cdot \mathbf{f_{in}}) \big).
    \vspace{-0.1cm}
\end{equation}

\vspace{-0.4cm}

\paragraph{Temporal Long-short Dependency Unit (T-LSU). }
For the temporal dimension, the video length is generally large. If we use the same method as spatial dimension, \ie, establishing dependencies by $T \times T$ weights for every pair of frames, it will consume too much calculation. Therefore, we leverage multiple 2D convolutional layers with kernels of different dilation coefficient $d$ and temporal size $t$ on the $C_{out}\times T \times N$ feature maps to learn the multi-scale long-short term dependencies (see Fig.~\ref{fig:comps} (c)). The T-LSU can be formulated as:
\begin{equation}\label{t-ldu}
    \mathbf{f_{out}}=\tlsu(\mathbf{f_{in}})= \conv_{2D (t\times1, d)}(\mathbf{f_{in}}).
    \vspace{-0.1cm}
\end{equation}

\vspace{-0.4cm}


\paragraph{Joint-bone Two-stream Fusion.}\label{jbf}
Inspired by the two-stream methods~\cite{2sagcn2019cvpr, Shi2019SkeletonBasedAR, liu2020disentangling}, we use a two-stream framework where a separate model with identical architecture is trained using the bone features initialized as vector differences of adjacent joints directed away from the body center. The $\softmax$ scores from the joint and bone models are summed to obtain final prediction scores.

\subsection{Design Strategy}
In this section, we present our design strategy that goes beyond GCNs by using a generic dependency matrix $\mathbf{W}_i$ (see Eq.~\ref{s-ldu1}) and the attention mechanism $\mathbf{A}_i$ to model the relations between joints in our unified formulation. 

\vspace{-0.3cm}

\paragraph{Dependency Matrix.} \label{dm}
For many human actions, the natural connectivity between joints are not the most appropriate to be used to extract features on skeletons (\eg, for ``drinking'', the connectivity between the head and the hand should be considered, but the original human topology does not include this connectivity). Hence, it is still an open question what kind of adjacency matrix can represent the optimal dependencies between joints for effective feature extraction. Recent works~\cite{Li_2019_CVPR, 2sagcn2019cvpr, liu2020disentangling} aim at optimizing the adjacency matrices to increase the receptive field of graph convolutions, by higher-order polynomials to make distant neighbors reachable~\cite{Li_2019_CVPR} or leveraging an attention mechanism to guide the learning process of the adjacency matrix~\cite{2sagcn2019cvpr,liu2020disentangling}. Specifically, they decompose the adjacency matrix into a certain number of subsets according to the distances between joints~\cite{liu2020disentangling} or according to the orientation of joints to the gravity (\ie, body center)~\cite{2sagcn2019cvpr}, so that each subset is learned individually by the self-attention. The learned feature maps are then aggregated together for the action classification. However, the number of subsets is constrained by the body structure. Moreover, we note that the manually pre-defined subsets of the adjacency matrix with prior knowledge (\ie, pre-defined body topology) are all sparse. At the initial learning stage, this spatial convolution relies on a graph-representation, while at the deeper stage, the relations coded within the adjacency matrix are no more sparse and the joint connections are represented by a complete-graph, which corresponds to a fully connected layer in the narrow sense. Finally, the dependencies converge to a sparse representation again, which is locally optimal but completely different from the original topological connectivity of the human body (see Fig.~\ref{fig:weights}). This motivates us, in this work, to revise the ``adjacency matrix'' by a generic ``dependency matrix'' that is prospectively initialized with a fully dense and uniform distribution (Eq.~\ref{s-ldu2}) to better reach the globally optimal representation. 

\vspace{-0.3cm}

\paragraph{Multi-head Aggregation.}
With our proposed initialization strategy, we can repeat the self-attention mechanism by leveraging multiple dependency matrices and sum the outputs to automatically aggregate the features focusing on different body joints (Eq.~\ref{s-ldu1}). As the number of attention maps (\ie, heads) $N$ is no longer limited by the human topology, we can use it as a flexible hyper-parameter to improve the model. In the ablation study (see Fig.~\ref{fig:weights} and Tab.~\ref{tab_ab1}), our insight has been verified. Overall, our design strategy makes the architecture more flexible, effective and generic, which facilitates the study of cross-domain transfer learning in this field for datasets using different joint distributions (see Fig.~\ref{fig:joints}).

\section{Posetics Skeleton Dataset}\label{sec:dataset}

In this section, we introduce Posetics, a novel large-scale pre-training dataset for skeleton-based action recognition (as illustrated in Fig.~\ref{figure:intro}). The Posetics dataset is created to study the transfer learning on skeleton-based action recognition. It contains 142,000 real-world video clips with the corresponding 2D and 3D poses including 17 body joints.
All video clips in Posetics dataset are filtered from Kinetics-400~\cite{Carreira_2017_CVPR}, to contain at least one human pose over $50\%$ of frames. 
\vspace{-0.3cm}
\paragraph{Motivation and Data Collection.} Recent skeleton-based action recognition methods on NTU-RGB+D~\cite{Shahroudy2016NTURA, NTU-120} can perform similarly or better compared to RGB-based methods. However, as laboratory indoor datasets may not contain occlusions, it is difficult to use such datasets to pre-train a generic model that can be transferred onto real-world videos, where skeleton data encounters a number of occlusions and truncations of the body. On the other hand, the accuracy based on skeleton data on the most popular real-world pre-training dataset, Kinetics~\cite{Carreira_2017_CVPR}, is still far below the accuracy on other datasets. The main problems are: (i) the skeleton data is hard to obtain by pose estimators as Kinetics is not human-centric. Human body may be missing or truncated by the image boundary in many frames. (ii) Many action categories are highly related to objects rather than human motion (\eg, ``making cakes'', ``making sushi'' and ``making pizza''). These make it difficult to effectively learn the human skeleton representation for recognizing actions. Hence, recent datasets~\cite{NTU-120, Yan2018SpatialTG} are unable to significantly boost the action recognition performance when applied to different datasets. In order to better study the
generalizability of skeleton-based models
in the real-world, we extract the pose (\ie, skeleton) data on Kinetics-400~\cite{Carreira_2017_CVPR} videos. Specifically, we compare the recent pose estimators and extract pose data from RGB videos through multiple pose estimation systems. Then we apply SSTA-PRS~\cite{Yang_2021_WACV}, a pose refinement system, for obtaining higher quality pose data in real-world videos. This system aggregates the poses of three off-the-shelf pose estimators~\cite{RogezWS18, OpenPose, fang2017rmpe}, as pseudo ground-truth and retrain LCRNet++~\cite{RogezWS18} to improve the estimation performance. Moreover, for the problem (i), we filter out the videos where no body detected, and for the problem (ii), we slightly and manually modify the video category labels of Kinetics-400, and place emphasis on relating verbs to poses. (\eg, For ``making cakes'', ``making sushi'' and ``making pizza'', we collectively chose the label ``making food'', whereas ``washing clothes'', ``washing feet'', and ``washing hair'' remain with the original labels). All in one, we organize 320 action categories for Posetics and this dataset can be more appropriately used for studying the real-world generalizability of skeleton-based action recognition models across datasets by transfer learning.

\section{Experiments and Analysis}
\begin{figure*}[t]
\begin{center}

\includegraphics[width=1.0\linewidth]{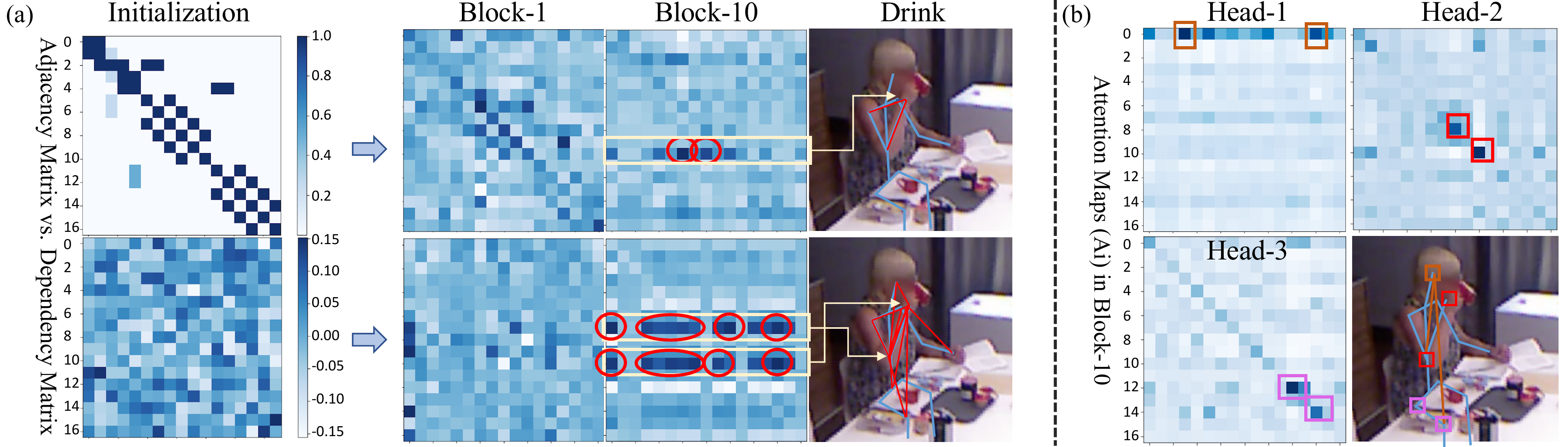}
\end{center}
\vspace{-0.4cm}
\caption{\small (a)~\textbf{Adaptive Adjacency Matrix~\cite{2sagcn2019cvpr} (top) vs. Dependency Matrix (bottom)} in different blocks for action "Drink" of Smarthome (right). They have different initial distributions. During training, the dependencies will become optimized representations, that are salient and more sparse in the deeper blocks, while our proposed matrix represents longer range dependencies (indicated by the red circles and red lines). (b)~\textbf{Multi-head attention maps} in Block-10. Similar to dependency matrices, attention maps are salient and sparse in the deep block. The different heads automatically learn the relationships between the different body joints (as shown in the boxes and lines with different colors) to process long-range dependencies between joints instead of using pre-defined adjacency matrices.}
\vspace{-0.2cm}
\label{fig:weights}
\end{figure*}

\subsection{Experimental Settings}
\paragraph{Overview.} Extensive experiments are conducted on 5 action classification datasets: \textbf{Toyota Smarthome (Smarthome)~\cite{Das_2019_ICCV}, Penn Action~\cite{penn}, NTU-RGB+D 60 (NTU-60)~\cite{Shahroudy2016NTURA}, NTU RGB+D 120 (NTU-120)~\cite{NTU-120}} and the proposed \textbf{Posetics}. See the supplementary material for datasets and implementation details pertaining to all experiments.
Firstly, we perform (i) exhaustive ablation study on Smarthome and NTU-60 without pre-training to verify the effectiveness of our proposed \textit{dependency matrix} and \textit{multi-head attention}. Then we (ii) demonstrate the impact of pre-training on Posetics by both linear evaluation and fine-tuning on Smarthome and Penn Action. (iii) We re-evaluate state-of-the-art models~\cite{Yan2018SpatialTG, 2sagcn2019cvpr, liu2020disentangling}, as well as our model on the proposed Posetics dataset (baselines are shown in Tab.~\ref{tab_sh}), proceed to provide an analysis on how much performance improves on target datasets: Smarthome, Penn Action, NTU-60 and NTU-120, after pre-training on Posetics. We demonstrate that our model generalizes well and benefits the most from pre-training. (iv) Final fine-tuned models are evaluated on all datasets to compare with the other state-of-the-art approaches for action recognition.
\vspace{-0.3cm}
\paragraph{Evaluation Protocols.}


For Posetics, we split the dataset into 131,268 training clips and 10,669 test clips. We use Top-1 and Top-5 accuracy as evaluation metrics~\cite{Yan2018SpatialTG}. With respect to real-word settings, 2D poses extracted from images and videos tend to be more accurate than 3D poses, which are more prone to noise. Therefore, we only use 2D data for evaluation and comparison of the models on Posetics. We note that for pre-training, both can be used, 2D and 3D data, in order to obtain different models that can be transferred to datasets with different skeleton data.
For the other datasets, we evaluate cross-subject (CS on Smarthome, NTU-60 and 120), cross-view (CV1 and CV2 on Smarthome and CV on NTU-60), cross-setup (CSet on NTU-120) protocols and the standard protocol (on Penn Action). Unless stated, we use 2D data on Smarthome and Penn Action, 3D data on NTU-60 and 120. 
\subsection{Ablation Study of UNIK}\label{ablation}

\paragraph{Impact of Dependency Matrix.}
 \begin{table}[t]
\centering
\begin{center}
\scalebox{0.92}{

\setlength{\tabcolsep}{2.6mm}{
\begin{tabular}{ l |c |c |c}
\hline
\multirow{3}*{\textbf{Datasets} (J)}
&\multicolumn{3}{c}{\textbf{Matrix}}  
\\

&\multicolumn{3}{c}{($N=3,\tau=1$)}  \\

\cline{2-4}
&F-AdjM & A-AdjM & DepM\\ 

\hline
\hline

\textbf{Smarthome CS (\%)}&50.4 &55.7&\textbf{58.5}\\

\textbf{NTU-60 CS (\%)}&84.3 &86.1&\textbf{87.3}\\

\hline
\end{tabular}}}
\end{center}
\vspace{-0.3cm}
\caption{Impact of Dependency Matrix on Smarthome CS and NTU-60 CS using joint (J) data only. F-AdjM: Fixed Adjacency Matrix (ST-GCN), A-AdjM: Adaptive Adjacency Matrix (AGCNs), DepM: Dependency Matrix (Ours).}
\vspace{-0.2cm}
\label{tab_ab1}
\end{table}

\begin{table}[t]
\centering
\begin{center}
\scalebox{0.92}{

\begin{tabular}{ l| c| c| c| c| c}
\hline
\multirow{3}*{\textbf{Datasets} (J)}

&\multicolumn{5}{c}{\textbf{\#Heads-$N$}} 
\\

&\multicolumn{5}{c}{($\tau=1$)} 
\\

\cline{2-6}
&0 & 1 & 3 & 6 & 12 \\ 

\hline
\hline

\textbf{Smarthome CS (\%)}&56.8& 58.1 &\textbf{58.5} &57.9 &58.1\\

\textbf{NTU-60 CS (\%)}&86.8& 87.0& 87.3& 87.1 & \textbf{88.0}\\

\hline
\end{tabular}}
\end{center}
\vspace{-0.3cm}
\caption{Impact of Multi-head Attention on Smarthome CS and NTU-60 CS using joint (J) data only.}
\vspace{-0.2cm}
\label{tab_ab2}
\end{table}

\begin{table}[t]
\centering
\begin{center}
\scalebox{0.92}{

\setlength{\tabcolsep}{1.5mm}{
\begin{tabular}{ l|c|c| c|c| c}
\hline
\multirow{3}*{\textbf{Datasets} (J)}
&\multicolumn{3}{c|}{\textbf{TW-$\tau$}}
&\multicolumn{2}{c}{\textbf{TD}}\\

&\multicolumn{3}{c|}{($N=3$)}
&\multicolumn{2}{c}{($N=3,\tau=1$)}\\

\cline{2-6}
& 1 & 2 & 6 & $\times$ &\checkmark\\ 

\hline
\hline

\textbf{Smarthome CS (\%)}&\textbf{58.5}&57.0&56.2&58.5&\textbf{58.9}\\

\textbf{NTU-60 CS (\%)}&87.3&87.0&\textbf{87.8}&87.3&\textbf{87.8}\\

\hline
\end{tabular}}}
\end{center}
\vspace{-0.3cm}
\caption{Impact of TW and TD on Smarthome CS and NTU-60 CS using joint (J) data only. TW: Temporal window size ($\tau$). TD: Temporal dilation.}
\vspace{-0.5cm}
\label{tab_ab3}
\end{table}

\begin{table*}[t]
\centering

\begin{center}

\scalebox{0.92}{
\setlength{\tabcolsep}{3.7mm}{
\begin{tabular}{  l c c c c c c c}

\hline
\multirow{2}*{\textbf{UNIK}}
&\multirow{2}*{\textbf{Pre-training}}
&\multicolumn{4}{c}{\textbf{Smarthome} (J+B)} & \multicolumn{2}{c}{\textbf{Penn Action} (J+B)}\\
&&\text{\#Params}& \text{CS(\%)} &\text{CV1(\%)} &\text{CV2(\%)}&\text{\#Params}& {\text{Accuracy (\%)}}\\
\hline
\hline
\rowcolor{mygray} Fine-tuning & Scratch &3.45M &63.1&22.9&61.2 & 3.45M & 94.0\\
\rowcolor{mygray} Fine-tuning & Posetics &3.45M &\textbf{64.3}&\textbf{36.1}&\textbf{65.2} & 3.45M & \textbf{97.9}\\

\hline
Linear classification & Scratch &7.97K &24.6 & 17.2&20.7 & 3.85K & 29.8\\

\text{Linear classification}& Posetics & 7.97K&\textbf{51.9}& \textbf{35.2}&\textbf{52.2} & 3.85K& \textbf{97.3}\\

\hline

\end{tabular}}}

\end{center}
\vspace{-0.3cm}
\caption{Mean per-class accuracy on Smarthome and Top-1 classification accuracy on Penn Action by \textit{Fine-tuning} (Backbone not fixed) and \textit{Linear classification} (Backbone fixed) for evaluation of extracted features by pre-training. ``J+B'': Joint and Bone two stream fusion.}
\vspace{-0.cm}
\label{tab_tr}
\end{table*}  

 \begin{table*}[t]
\centering

\begin{center}

\scalebox{0.92}{
\begin{tabular}{  l c c c c c c c c c}

\hline
\multirow{2}*{\textbf{Methods}}
&\multirow{2}*{\textbf{Pre-training}}
&\multicolumn{3}{c}{\textbf{Smarthome} (J)}
&\multicolumn{1}{c}{\textbf{Penn Action} (J)}
&\multicolumn{2}{c}{\textbf{*NTU-60} (J+B)}
&\multicolumn{2}{c}{\textbf{*NTU-120} (J+B)}\\

& & CS (\%) & CV1 (\%) & CV2 (\%) & Top-1 Acc. (\%) & CS (\%) & CV (\%)& CS (\%) & CSet (\%)\\

\hline
\hline

2s-AGCN~\cite{2sagcn2019cvpr}& Scratch &55.7 & 21.6 & 53.3 & 89.5  & 84.2 & 93.0 & 78.2&82.9 \\
MS-G3D~\cite{liu2020disentangling}& Scratch &55.9 & 17.4 & 56.7 & \text{88.5}& \textbf{86.0} & \textbf{94.1} & \textbf{80.2}& \textbf{86.1}\\
\text{UNIK (Ours)} & Scratch &\textbf{58.9} &\textbf{21.9} & \textbf{58.7} &\textbf{90.1} & 85.1& \text{93.6}& 79.1& 83.5 \\
\hline
2s-AGCN~\cite{2sagcn2019cvpr} & Posetics & 58.8& 32.2 & 57.9 &96.4& 85.8& 93.4& 79.7&85.0\\
MS-G3D~\cite{liu2020disentangling} & Posetics & {59.1} &26.6 & 60.1 &92.2& 86.2& 94.1 & \text{80.6} & \text{86.4} \\
\text{UNIK (Ours)} & Posetics & \textbf{\color{blue}62.1} & \textbf{\color{blue}33.4} & \textbf{\color{blue}63.6} & \textbf{\color{blue}97.2} & \textbf{\color{blue}86.8} &\textbf{\color{blue}94.4}& {\textbf{\color{blue}80.8}}& {\textbf{\color{blue}86.5}} \\
\hline

\end{tabular}}

\end{center}
\vspace{-0.3cm}
\caption{Generalizability study of state-of-the-art by comparing the impact of transfer learning on Smarthome, Penn Action, NTU-60 and 120 datasets. The {\color{blue}blue} values indicate the best generalizabilities that can take the most advantage of pre-training on Posetics. ``*'' indicates that we only use 17 main joints adapted to the pre-trained model on Posetics.}

\vspace{-0.4cm}
\label{tab_gen}
\end{table*}  


Here we compare the dependency matrices with the adaptive adjacency matrices. In order to verify our analysis in Sec.~\ref{dm}, we visualize the adjacency matrices~\cite{2sagcn2019cvpr} before and after learning. As shown in Fig.~\ref{fig:weights} (a)-top, we find that the previous learned graph~\cite{2sagcn2019cvpr} becomes a complete-graph, whose relationships are represented by weights that are well distributed over the feature maps. In contrast, our method is able to explore longer range dependencies, while being based on a dependency matrix with self-attention, which freely searches for dependencies of the skeleton from the beginning without graph-representation (see Fig.~\ref{fig:weights} (a)-bottom). Quantitatively, results in Tab.~\ref{tab_ab1} show the effectiveness of the Dependency Matrix. Overall, we conclude that, both our method and AGCN-based methods are fully connected layers with different initialization strategies and attention mechanisms in the spatial dimension, both are better than using a fixed graph~\cite{Yan2018SpatialTG}. It becomes evident that for skeleton-based tasks, where the number of nodes (\ie, spatial body joints) is not large, multi-head attention based dependency matrix learning along with temporal convolutions can be a more generic and effective way to learn spatio-temporal dependencies compared with graph convolution.
\vspace{-0.35cm}

\paragraph{Impact of Multi-head Attention.}

In this section, we visualize the multi-head attention maps and analyze the impact of the number of heads $N$ for UNIK with $N=1, 3, 6, 12$. As shown in Fig.~\ref{fig:weights}, our multi-head aggregation mechanism can automatically learn the relationships between different positions of body joints by conducting the spatial processing (see Eq.~\ref{s-ldu2}) using the unified dependency matrices with a uniform initialization. 
Quantitative results in Tab.~\ref{tab_ab2} show that obtaining a correct number of heads $N$ is instrumental in improving the accuracy in a given dataset, but weakens the generalization ability across datasets (\eg, the model benefits predominantly from $N=12$ for NTU-60, and $N=3$ for Smarthome). 
Consequently, we set $N=3$ as a unified setting for all experiments and all datasets in order to balance the efficiency and performance of the model, as well as the generalization ability. 




\begin{table*}[t]
\centering

\begin{center}

\scalebox{0.92}{
\setlength{\tabcolsep}{2.1mm}{
\begin{tabular}{  l c c c c c c c c c}

\hline
\multirow{2}*{\textbf{Methods}}
&\multirow{2}*{\textbf{RGB}}
&\multirow{2}*{\textbf{Pose}}

&\multirow{2}*{\textbf{Pre-training}}
&\multicolumn{2}{c}{\textbf{Posetics}}
&\multicolumn{3}{c}{\textbf{Smarthome}}
&\textbf{Penn Action}\\

& & & &\text{Top-1(\%)} &\text{Top-5(\%)}&\text{CS(\%)} &\text{CV1(\%)} &\text{CV2(\%)} &\text{Accuracy(\%)}\\
\hline
\hline
\rowcolor{mygray} I3D~\cite{Carreira_2017_CVPR}&\checkmark&&Kinetics-400&\color{blue}53.5 &\color{blue}78.0 &53.4 &34.9& 45.1&-\\
\rowcolor{mygray} AssembleNet++~\cite{Ryoo2020AssembleNetAM}&\checkmark&&Kinetics-400&-&-& \text{\color{blue}63.6} &-& -&-\\  
\rowcolor{mygray} NPL~\cite{unseenview} &\checkmark&&Kinetics-400&-&-& - &39.6& \text{\color{blue}54.6}&-\\  
\rowcolor{mygray} Separable STA~\cite{Das_2019_ICCV}& \checkmark&\checkmark&Kinetics-400&-& -&54.2& 35.2& 50.3&-\\
\rowcolor{mygray} VPN~\cite{das2020vpn} &\checkmark &\checkmark&Kinetics-400&-&-&\text{60.8} &\text{\color{blue}43.8} &\text{53.5}&-\\
\rowcolor{mygray} Multi-task~\cite{multitask} &\textbf{\checkmark}&\textbf{\checkmark}&Scratch&- &-&-&-&-& \text{\color{blue}97.4}\\

\hline

 LSTM~\cite{Mahasseni2016RegularizingLS}&&\textbf{\checkmark}&Scratch&-&-& 42.5 &13.4& 17.2&-\\
ST-GCN~\cite{Yan2018SpatialTG}&&\textbf{\checkmark}&Scratch& 43.3&67.3& \text{53.8} & 15.5& 51.1& 89.6\\
2s-AGCN~\cite{2sagcn2019cvpr}&&\textbf{\checkmark}&Scratch& 47.0&70.8&60.9 & \text{22.5} & 53.5 & 93.1\\
MS-G3D Net~\cite{liu2020disentangling}&&\textbf{\checkmark}&Scratch&47.1&70.0 &61.1 & 17.5 & 59.4& 92.7 \\
\textbf{UNIK (Ours)}&&\text{\checkmark}&Scratch& \textbf{47.6} &\textbf{71.3} &\textbf{63.1}& \textbf{22.9} &\textbf{61.2}&\textbf{94.0}\\
\hline

Pr-ViPE~\cite{sun2020viewinvariant} &&\textbf{\checkmark}&Human3.6M&-&-&-&-&-&97.5\\
\textbf{UNIK (Ours)}&& \text{\checkmark}&Posetics(Ours)& -&-&\textbf{64.3}& \textbf{36.1} &\textbf{65.0}&\textbf{97.9}\\

\hline

\end{tabular}}}

\end{center}
\vspace{-0.3cm}
\caption{Comparison with state-of-the-art methods on the Posetics, Toyota Smarthome and Penn Action dataset. The best results using RGB data are marked in {\color{blue}blue} for reference.}
\vspace{-0.1cm}
\label{tab_sh}
\end{table*}

\begin{figure*}[t]
\begin{center}

\includegraphics[width=1.0\linewidth]{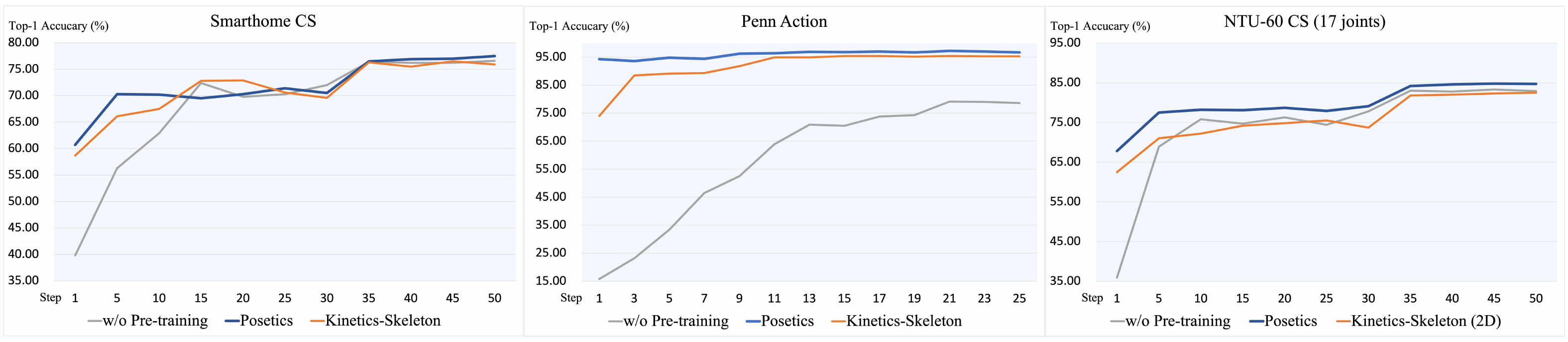}
\end{center}
\vspace{-0.3cm}
\caption{\small Validation accuracy with the training steps on Smarthome, Penn Action and NTU-60 datasets for demonstrating the impact of Pre-training on Posetics. Pre-training on Kinetics-Skeleton~\cite{Yan2018SpatialTG} (\ie, 2D Openpose on Kinetics-400) is shown for reference.}
\vspace{-0.4cm}
\label{fig:curve}
\end{figure*}

\vspace{-0.35cm}

\paragraph{Other Ablations.} For further analysis, 
results in Tab.~\ref{tab_ab3} also show that (i) similar to~\cite{liu2020disentangling}, the size of the sliding window (see~\ref{ld-learning}) $\tau$ can help to improve the performance, however weakening the generalizability of the model. (ii) Temporal dilated convolution contributes to minor boosts.


\subsection{Impact of Pre-training.}\label{exp:pretraining}


\paragraph{Transfer Learning.}
In this section, we evaluate the features of skeletons extracted by UNIK pre-trained on Posetics by transfer learning. For linear classification, we freeze the UNIK backbone pre-trained on Posetics, then retrain linear classifiers on smaller benchmarks: Smarthome and Penn Action. The results in Tab.~\ref{tab_tr} demonstrate the effectiveness of transfer learning with fewer parameters compared with classification from scratch. Fine-tuning (\ie, UNIK backbone not fixed) results are also shown. In addition, we visualize the training curve for the Top-1 accuracy with training steps during fine-tuning (see Fig.~\ref{fig:curve}). From the curves, we deduce that at the beginning of training steps, the pre-training has a significant boost for all transferred datasets. This suggests that the weights of the model are well pre-trained on Posetics, providing a strong transfer ability \ie, pre-trained on Posetics is generic and can be used for extracting features of skeleton sequences. 

\vspace{-0.35cm}

\paragraph{Generalizability Study.}
In this section, we report the classification results on all the four target datasets to demonstrate the impact of pre-training and compare the generalization capacities (\ie, benefits by fine-tuning compared to training from scratch) with state-of-the-art methods. Specifically, we pre-train respectively \cite{2sagcn2019cvpr, liu2020disentangling} and our proposed UNIK in a unified setting ($N=3, K=10, \tau=1$). Note that for pre-training GCN-based models~\cite{2sagcn2019cvpr, liu2020disentangling}, we need to
manually calibrate the different human topological structures in different datasets to keep the pre-defined graphs unified. Note that unless otherwise stated, we use the consistent skeleton data (2D on Smarthome, Penn Action and 3D on NTU-60, 120), number of joints (17 main joints) for fair comparison of all models. On NTU-60 and 120, we use both joint (J) and bone (B) data to compare the full models with two-stream fusion.  
The results suggest that pre-training consistently boosts all models, see Tab.~\ref{tab_gen}, in particular, small benchmarks (\eg, Smarthome CV and Penn Action with $5\%\sim12\%$ improvement), as we do not have sufficiently large training data. Previous work~\cite{liu2020disentangling} has a weak transfer capacity, due to the dataset-specific model settings (\eg, the number of GCN scales and G3D scales) not always being able to adapt to the transferred datasets. On NTU-60, we take the main 17 joints for fine-tuning as we estimate and refine the main 17 joints on Posetics, and our pre-trained model outperforms state-of-the-art model~\cite{liu2020disentangling}. Therefore, we conclude that our pre-trained model is the most generic-applicable especially for real-world scenarios. 



\subsection{Comparison with State-of-the-art}

We compare our full model (\ie, Joint+Bone fusion) with and without pre-training to state-of-the-art methods, reporting results in Tab.~\ref{tab_sh} (Posetics, Smarthome and Penn Action). Note that for fair comparison, we use the same skeleton data (2D and 17 joints) for all models. For real-world benchmarks using estimated skeleton data (\eg, Posetics, Smarthome and Penn Action), our model without pre-training outperforms all state-of-the-art methods~\cite{Mahasseni2016RegularizingLS, Yan2018SpatialTG, 2sagcn2019cvpr, liu2020disentangling} in skeleton (\ie, pose) stream and with pre-training, outperforms the embedding-based method~\cite{sun2020viewinvariant} that pre-trained on Human3.6M~\cite{h36m_pami}. On NTU-60 and 120 (see Tab.~\ref{tab_gen}), we compare the most impressive two-stream graph-based methods~\cite{2sagcn2019cvpr, liu2020disentangling} and our model performs competitively without pre-training. We argue that, we simplify our model as generically as possible without data-specific settings, which can improve the performance but weaken the transfer behavior (\eg, the setting of $N$ and $\tau$). Subsequently, we further compare RGB-based methods~\cite{Carreira_2017_CVPR, Das_2019_ICCV, Ryoo2020AssembleNetAM, unseenview, das2020vpn, multitask} for reference, that can be pre-trained on Kinetics-400~\cite{Carreira_2017_CVPR}. It suggests that previous skeleton-based methods~\cite{Mahasseni2016RegularizingLS, Yan2018SpatialTG, 2sagcn2019cvpr, liu2020disentangling} without leveraging the pre-training are limited by the poor generalizability and the paucity of pre-training data. In contrast, our proposed framework, UNIK with pre-training on the Posetics dataset, outperforms state-of-the-art using RGB and even both RGB and pose data on the downstream tasks (\eg, Smarthoma and Penn Action). 


\section{Conclusion}
In this paper, we 
have proposed UNIK, a unified framework for real-world skeleton-based action recognition. Our experimental analysis shows that UNIK is effective and has a strong generalization ability to transfer across datasets. 
In addition, we have introduced Posetics, a large-scale real-world skeleton-based action recognition dataset featuring high quality skeleton annotations. Our experimental results demonstrate that pre-training on Posetics improves performance of the action recognition approaches. 
Future work involves an analysis of our framework for additional tasks involving skeleton sequences (\eg, 2D-to-3D pose estimation).




{\small
\balance{
\bibliographystyle{ieee_fullname}
\bibliography{egbib}
}

\clearpage
\appendix
  \renewcommand{\appendixname}{Appendix~\Alph{section}}
\section*{{Appendix}}

In this supplementary material we provide additional details \wrt, our experimental analysis provided in the main paper. In section~\ref{ex_detail}, we provide details pertaining to the datasets, the implementation of our framework. In section~\ref{detail-posetics}, we provide more details on the proposed Posetics dataset including the comparison to other related datasets. 

\section{Experimental Details}\label{ex_detail}
\subsection{Datasets}
\paragraph{Toyota Smarthome.}
Toyota Smarthome~\cite{Das_2019_ICCV} (Smarthome) is a real-world dataset for daily living action classification, recorded in an apartment, where 18 older subjects carry out tasks of daily living during a day. The dataset contains 16,115 videos of 31 action classes, and the videos are taken from 7 different camera viewpoints. All actions are performed in a natural way without strong prior instructions. It provides RGB videos and two versions of skeleton data, which is extracted either from LCRNet++~\cite{RogezWS18} (v1) or from SSTA-PRS~\cite{Yang_2021_WACV} (v2). In this work, we use the skeleton-v2 for all experiments and comparisons in skeleton stream. Unless stated, we only use 2D data for the experiments. For the evaluation on this dataset, we follow the cross-subject (CS) and cross-view (CV1 and CV2) evaluation protocols.\par
\vspace{-0.3cm}

\paragraph{Penn Action.}
Penn Action dataset~\cite{penn} contains 2,326 video sequences of 15 different actions and human joint annotations for each sequence. Given that annotated skeletons have a large number of missing joints due to occlusions and truncations, we use LCRNet++~\cite{RogezWS18} to obtain the 2D skeletons for experiments. We report Top-1 accuracy following the standard train-test split.

\vspace{-0.3cm}

\paragraph{NTU-RGB+D 60.}
NTU-RGB+D 60~\cite{Shahroudy2016NTURA} (NTU-60) is a large-scale multi-modality dataset which consists of 56,880 sequences of high-quality 2D and 3D skeletons with 25 joints, associated with depth maps, RGB and IR frames captured by the Microsoft Kinect v2 sensor. We only use sequences of 3D skeletons in this work and we follow the cross-subject (CS) and cross-view (CV) evaluation protocols.

\vspace{-0.3cm}

\paragraph{NTU-RGB+D 120.}
NTU-RGB+D 120~\cite{NTU-120} (NTU-120) dataset extends the number of action classes and videos of NTU-RGB+D 60 to 120 classes 114,480 videos. Similarly, we use only sequences of 3D skeletons and we follow the cross-subject (CS) and cross-set (CSet) evaluation protocols.

\subsection{Implementation Details}

\paragraph{Implementation of UNIK.}
Unless otherwise stated in the ablation study, all UNIK models have $N = 3, \tau=1$ for S-LSU, and $t=9$, $d=1, 3, 3, 3, 3, 1, 1, 1, 1, 1$, in each block respectively for T-LSU. We use SGD for training with momentum 0.9, an initial learning rate of 0.1 for 50, 30, 50, 60, and 65 epochs with step LR decay with a factor of 0.1 at epochs \{30, 40\}, \{10, 20\}, \{30, 40\}, \{30, 50\}, and \{45, 55\} for Smarthome, Penn Action, NTU-60, NTU-120, and Posetics, respectively. Weight decay is set to 0.0001 for final models. For NTU-60 and 120, all skeleton sequences are padded to 300 frames by replaying the actions. For Smarthome, Penn Action, Posetics, we randomly choose 400, 150, 150 frames respectively for each training epoch and all frames for test. 2D and 3D inputs are pre-processed with normalization and centering following~\cite{pavllo:videopose3d:2019}, \cite{2sagcn2019cvpr} respectively. As we have both 2D and 3D skeleton data on Posetics, we pre-train two models for transferring to benchmarks with different types of skeleton data. Note that for ablation study of UNIK (Sec.~\ref{ablation}), we train all models from scratch, without pre-training.

\vspace{-0.3cm}

\paragraph{Number of Joints.}
SSTA-PRS~\cite{Yang_2021_WACV} and LCRNet++~\cite{RogezWS18} provide 13 joints of the main body. We add "hip", "chest", "neck" and "nose" by interpolation and obtain 17 joints for all experiments of real-world datasets (\ie, Posetics, Smarthome, Penn Action). On NTU-60 and 120, we use 3D Kinect skeleton data with 25 joints for ablation study of UNIK (Sec.~\ref{ablation}) while 17 main body joints for generalizability study (Sec.~\ref{exp:pretraining}) to adapt to the pre-trained model on Posetics.

\section{Details of Posetics Dataset}\label{detail-posetics}
In this section, we first review the recent skeleton-based action recognition datasets and then compare the most impressive ones with Posetics by (i) the benefit of pre-training for smaller datasets and (ii) skeleton quality obtained by different pose estimators.

\vspace{-0.3cm}

\paragraph{Review of Skeleton Datasets.}

Tab.~\ref{tab_datasets} shows an overview of pertinent skeleton-based action recognition datasets, which we proceed to describe. To evaluate methods in 3D human action recognition, \cite{Shahroudy2016NTURA, NTU-120, ucla_2014_CVPR} were recorded in laboratory conditions, where acquired actions were performed by actors under strict guidance. In contrast, \cite{Das_2019_ICCV, penn, Yan2018SpatialTG, skeletics152, uav} aimed to explore real-world action recognition using estimated or handcrafted annotated skeleton data. As the largest real-world dataset, Kinetics-Skeleton~\cite{Yan2018SpatialTG} provides poses extracted using OpenPose~\cite{OpenPose}. GCNs were applied on the real-world videos~\cite{Carreira_2017_CVPR} using pseudo 3D data \ie, 2D data and confidence~\cite{OpenPose}. However, the pose quality is limited due to occlusions and truncations. Skeletics-152~\cite{skeletics152} addressed this issue by applying VIBE~\cite{vibe} to obtain higher-quality skeleton data from Kinetics-700~\cite{k700} and manually scaled down the dataset by omitting the object-oriented action categories. However, there are still videos with missing skeletons. Deviating from the above, the higher-quality skeleton data in our dataset is calculated using pose refinement method~\cite{Yang_2021_WACV} which aims at real-world pose estimation~\cite{Das_2019_ICCV}. Instead of omitting the samples in object-oriented action categories, we merge action categories that are incompatible with skeleton-based action recognition to keep the scale and we filter out the non-skeleton videos. Our large-scale dataset can be more effectively used for pre-training and transferring onto other simulated~\cite{Shahroudy2016NTURA, NTU-120} and real-world~\cite{Das_2019_ICCV, penn} scenarios. Besides, our dataset can be used for human action video generation~\cite{WANG_2020_WACV,Wang_2020_CVPR,wang2021inmodegan,Wang_2018_ECCV_Workshops,Chan_2019_ICCV} towards augmenting existing datasets. \par

\vspace{-0.3cm}

\begin{table*}[!th]
\centering

\begin{center}

\scalebox{0.75}{
\setlength{\tabcolsep}{3.8mm}{
\begin{tabular}{  l c c c c c c c c c}
\hline

\textbf{Dataset}& \textbf{RW}&\textbf{2D }&\textbf{3D} &\textbf{\#Videos}&\textbf{\#Classes}&\textbf{\#Joints}&\textbf{Skeleton data}&\textbf{Skeleton quality} &\textbf{Year}\\ 
\hline
\hline

Human3.6M~\cite{h36m_pami}&$\times$ &\checkmark &\checkmark &209 &15& 24& Motion Capture system &High&2014\\

NTU-RGB+D 60~\cite{Shahroudy2016NTURA} &$\times$ &\checkmark &\checkmark &56,880 &60 & 25 & Kinect v2.0 &High& 2016\\ 
NTU-RGB+D 120~\cite{NTU-120} &$\times$ &\checkmark &\checkmark &114,480 &120 &25  &  Kinect v2.0 &High&2019\\ 

Penn Action~\cite{penn}&\checkmark &\checkmark &$\times$ &2,326 &15 & 13  & Handcrafted annotation& Medium&2013\\ 
UAV-Human~\cite{uav}
&\checkmark &\checkmark &$\times$ & 21,224& 155 & 17  & AlphaPose~\cite{fang2017rmpe}& Medium& 2021
\\
Kinetics-Skeleton~\cite{Yan2018SpatialTG}
&\checkmark &\checkmark &$\times$ &260,232 &400 & 18  & OpenPose~\cite{OpenPose}& Low& 2018
\\ 

Toyota Smarthome~\cite{Das_2019_ICCV} 
&\checkmark &\checkmark &\checkmark &16,115 &31 & 13  & LCRNet++~\cite{Yang_2021_WACV}& Low&2019
\\

Skeletics-152~\cite{skeletics152}
&\checkmark &\checkmark &\checkmark &125,621 &152 & 25  & VIBE~\cite{vibe}& High& 2021
\\ 

\hline

\textbf{Posetics (Ours)} &\checkmark &\checkmark &\checkmark &142,000 &320 & 25  & SSTA-PRS~\cite{Yang_2021_WACV}+Intrpl. &High& 2021\\ 
\hline

\end{tabular}
}}

\end{center}
\vspace{-0.1cm}
\caption{A survey of recent datasets for skeleton-based action recognition. ``RW'': Real-world. ``Intrpl.'': Interpolation.}
\vspace{-0.cm}
\label{tab_datasets}
\end{table*}

\paragraph{Comparison of Pre-training.}\label{com_datasets}
In this section, we compare Posetics with NTU-120~\cite{NTU-120} and Kinetics-Skeleton~\cite{Yan2018SpatialTG} datasets by fine-tuning performances after pre-training. We note that we only have 2D skeletons on Kinetics-Skeleton for pre-training. Consequently, we use 2D data of NTU-60 for fine-tuning. Results in Tab.~\ref{tab_comdatasets} demonstrate the effectiveness of our Posetics dataset (\ie, pre-training on Posetics boosts the most on target datasets).

\begin{table}[t]
\centering

\begin{center}

\scalebox{0.63}{
\setlength{\tabcolsep}{0.8mm}{
\begin{tabular}{  l c|| c c c|| c|| c }

\hline
\multirow{2}*{\textbf{Methods}}
&\multirow{2}*{\textbf{Pre-training}}
&\multicolumn{3}{c||}{\textbf{Smarthome }}
&\multicolumn{1}{c||}{\textbf{Penn Action}}
&\multicolumn{1}{c}{\textbf{*NTU-60}} \\

&& CS (\%) & CV1 (\%) & CV2 (\%) & Top-1 Acc. (\%) & CS(\%)\\

\hline
\hline

\text{UNIK-J} & NTU-RGB+D 120 &59.2 &28.3& 59.7 &91.7&-\\
\text{UNIK-J} & Kinetics-Skeleton &58.9& 29.5 &60.6& 95.4 & $\dagger$82.5\\
\textbf{UNIK-J} & Posetics (Ours) &\textbf{62.1} & \textbf{33.4} & \textbf{63.6} & \textbf{97.2} & \textbf{85.3} \\
\hline
\text{UNIK-B} & Posetics (Ours) &61.1 & 31.3 & 62.5 & 97.4 & 84.9  \\
\textbf{UNIK-J\&B} & Posetics (Ours)&\textbf{64.3} & \textbf{36.1} & \textbf{65.0} & \textbf{97.9}& \textbf{86.8} \\
\hline

\end{tabular}}}

\end{center}
\vspace{-0.1cm}
\caption{Comparison of datasets by pre-training (top) and impact of two-stream fusion (bottom). ``J''/``B'': Joint/Bone stream. ``$\dagger$'': The input data (2D) is different from other competitors (3D) on NTU-60 due to the lack of 3D data on Kinetics-Skeleton. ``*'': We only use 17 main joints.}

\vspace{-0.cm}
\label{tab_comdatasets}
\end{table}  

\vspace{-0.3cm}

\paragraph{Comparison of Pose Estimators.}
In this section, we compare Posetics with Kinetics-Skeleton~\cite{Yan2018SpatialTG} and Skeletics-152~\cite{skeletics152} by the pose (\ie, skeleton) quality. While \cite{Yan2018SpatialTG}, \cite{skeletics152} use OpenPose~\cite{OpenPose} and VIBE~\cite{vibe} respectively to estimate poses, Posetics uses SSTA-PRS~\cite{Yang_2021_WACV} that integrates the advantages of three pose estimators including OpenPose~\cite{OpenPose}. Hence, Posetics has higher quality poses, in particular in cases of occlusions and truncations (see Fig.~\ref{fig:pose1} and Fig.~\ref{fig:pose2} for qualitative comparison). For quantitative comparison, we lack ground-truth poses, and hence we indirectly evaluate the quality of poses through the performance of action recognition. Towards this, we use all clips of Posetics and Smarthome with different 2D pose data for action recognition. Experimental results in Tab.~\ref{tab_pose} show that the performance using SSTA-PRS poses is higher than that using other poses.

\begin{table}[ht]
\centering
\begin{center}
\scalebox{0.63}{

\setlength{\tabcolsep}{5mm}{
\begin{tabular}{ l ||c c|| c }
\hline
\multirow{2}*{\textbf{UNIK}}  & \multicolumn{2}{c||}{\textbf{Posetics}(JB)}& \multicolumn{1}{c}{\textbf{Smarthome}(J)} \\

& \text{Top-1(\%)} &\text{Top-5(\%)} & \text{CS(\%)}\\
\hline
\hline
OpenPose~\cite{OpenPose}&45.9 & 69.5&-\\
VIBE-Pose~\cite{vibe}&- & -&42.5\\
VIBE-Mesh~\cite{vibe}&-& -& 43.2\\
\text{SSTA-PRS}~\cite{Yang_2021_WACV} &\textbf{47.6} &\textbf{71.3}& \textbf{58.9} \\

\hline
\end{tabular}}}
\end{center}
\vspace{-0.1cm}
\caption{Classification accuracy of UNIK using different poses on Posetics and Smarthome.}
\vspace{-0.cm}
\label{tab_pose}
\end{table}

\begin{figure*}[t]
\begin{center}

\includegraphics[width=0.85\linewidth]{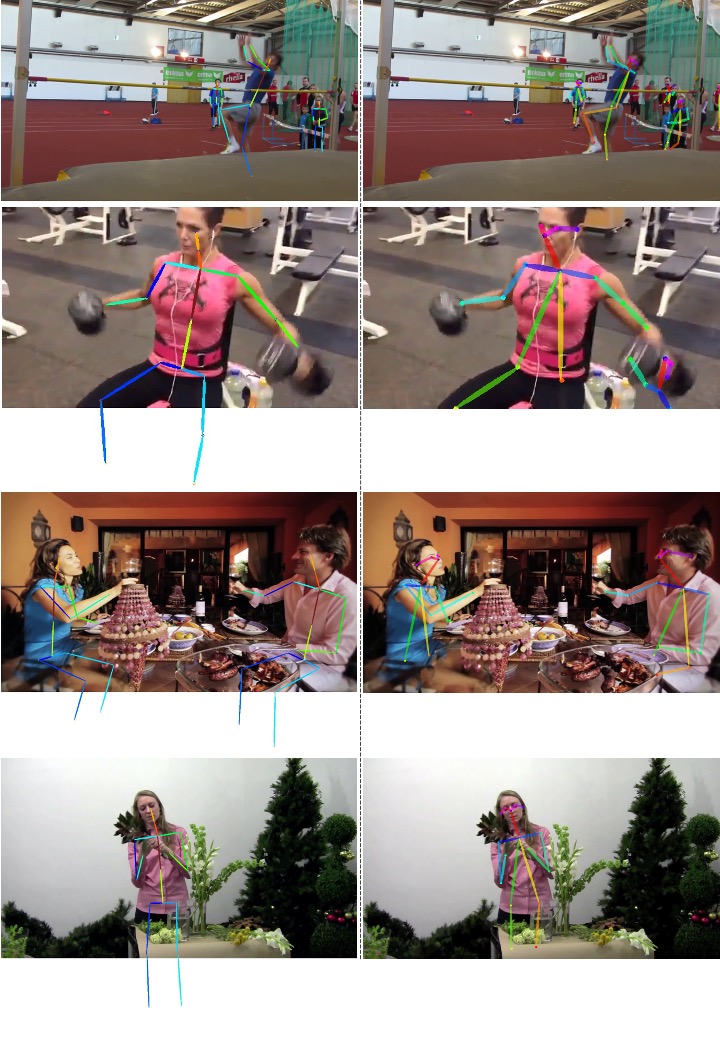}
\end{center}
\vspace{-0.65cm}
\caption{\textbf{Visualization of pose data-I} in Posetics (left) and Kinetics-Skeleton (right) in the case of occlusions and truncations.}
\vspace{-0.0cm}
\label{fig:pose1}
\end{figure*}

\begin{figure*}[t]
\begin{center}

\includegraphics[width=0.8\linewidth]{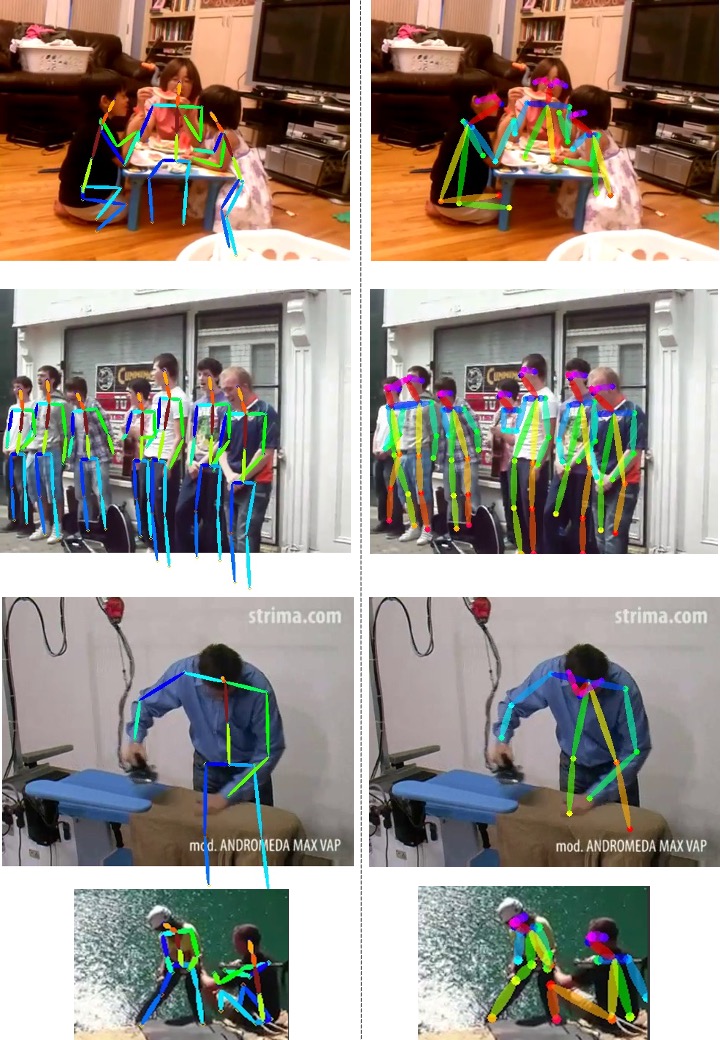}
\end{center}
\vspace{-0.45cm}
\caption{\textbf{Visualization of pose data-II} in Posetics (left) and Kinetics-Skeleton (right) in the case of occlusions and truncations.}
\vspace{-0.0cm}
\label{fig:pose2}
\end{figure*}

}
\end{document}